\documentclass[a4paper, 10pt, conference]{ieeeconf}      % Use this line for a4
\usepackage{FG2027}

\FGfinalcopy % *** Uncomment this line for the final submission

\IEEEoverridecommandlockouts                              % This command is only
\usepackage{graphics} % for pdf, bitmapped graphics files
\usepackage{graphicx}
\usepackage{amsmath} % assumes amsmath package installed
\usepackage{amssymb}  % assumes amsmath package installed
\usepackage{booktabs}
\usepackage{multirow}

\usepackage{pifont}  % Required for dingbat symbols

\newcommand{\cmark}{\ding{51}} % Defines \cmark as a checkmark
\newcommand{\xmark}{\ding{55}} % Optional: Defines \xmark as a cross

\def\FGPaperID{319} % *** Enter the FG2027 Paper ID here

\title{\LARGE \bf
Printed and Scanned: Exposing the Fragility of Single-Image \\ Morphing Attack Detectors under Print-Scan Degradation
}

\author{\parbox{16cm}{\centering
    {\large Diogo J. Paulo$^{1,2,3}$, Hugo Proença$^{1,3}$ and João C. Neves$^{1,2}$}\\
    {\normalsize
    $^1$ University of Beira Interior\quad
    $^2$ NOVA LINCS\quad
    $^3$ IT: Instituto de Telecomunicações}}
}

\begin{document}

\ifFGfinal
\thispagestyle{empty}
\pagestyle{empty}
\else
\author{Anonymous FG2027 submission\\ Paper ID \FGPaperID \\}
\pagestyle{plain}
\fi
\maketitle

%%%%%%%%%%%%%%%%%%%%%%%%%%%%%%%%%%%%%%%%%%%%%%%%%%%%%%%%%%%%%%%%%%%%%%%%%%%%%%%%
\begin{abstract}
Face morphing attacks, in which the biometric traits of two or more identities are blended into a single face image, pose a severe threat to identity document issuance and automated border control. Consequently, a large body of work has proposed Single-image Morphing Attack Detection (S-MAD) algorithms and benchmarked them on datasets such as SYN-MAD22. However, the majority of these benchmarks operate exclusively in the digital domain, which contrasts with the real-world passport issuance pipeline in some countries, which still requires an applicant-supplied photograph to be printed and subsequently re-scanned before it is embedded in the identity document or fed into an automated verification system. This print-scan step introduces halftoning, resampling, color-space, and sensor noise that are absent from digital pipelines, and may mask the high-frequency artifacts that S-MAD models rely on. In this work, we introduce a print-scan version of the SYN-MAD22 dataset, obtained by printing and re-scanning all bona fide and morphed images at 300 and 600 DPI, and we use it to evaluate eight state-of-the-art S-MAD models trained exclusively on digital data. We additionally repeat the acquisition with a second scanner to assess cross-device robustness. Our results show that detectors that achieve near state-of-the-art performance in the digital domain suffer a more than threefold increase in average equal error rate, with the same qualitative degradation observed across both scanning devices, exposing a critical, previously under-reported gap between digital-benchmark performance and robustness to print-scan degradation.
\end{abstract}

 \begin{figure}[ht]
    \centering
    % \fbox{\rule{0pt}{4.2cm}\rule{\linewidth}{0pt}}
    \includegraphics[width=.9\linewidth]{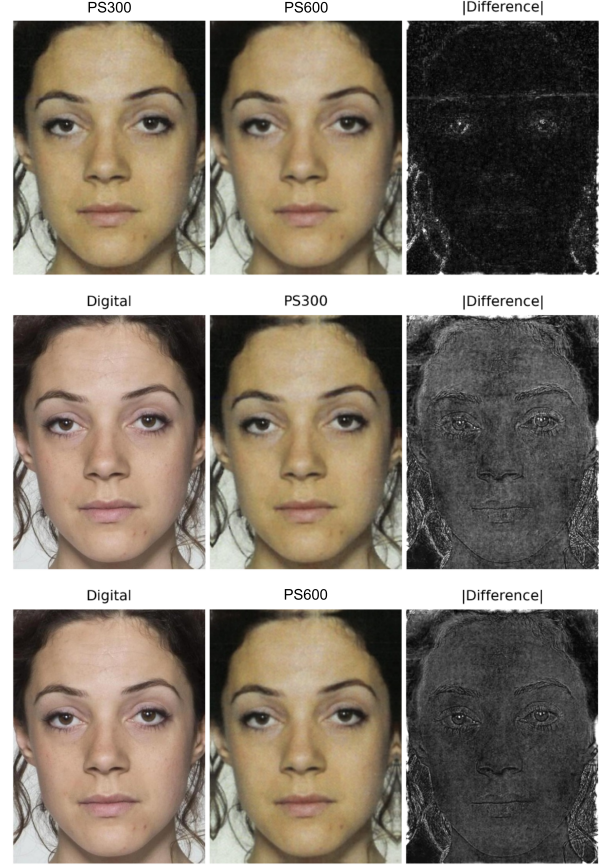}
    \caption{\textbf{Visual comparison of the digital and print-scanned versions of a representative SYN-MAD22 image.} Absolute difference maps reveal spatial changes in facial texture, edges, and color between the digital and print-scanned images, whereas the differences between the two scanning resolutions are comparatively small.}
    \label{fig:teaser}
\end{figure}

%%%%%%%%%%%%%%%%%%%%%%%%%%%%%%%%%%%%%%%%%%%%%%%%%%%%%%%%%%%%%%%%%%%%%%%%%%%%%%%%
\section{Introduction}
\label{sec:intro}
In a morphing attack, the facial images of two or more individuals are digitally combined into a single image that is perceptually similar to both contributors, such that a face verification system or a human examiner may accept either individual as a match to the morphed image~\cite{ferrara2014magic}. When such a morphed photograph is submitted during the application for an identity document (e.g., a passport), the resulting document can subsequently be used by any of the contributing accomplices to cross automated border control, enabling identity-sharing between a look-alike accomplice and a criminal who would otherwise not be able to obtain the document. This attack vector, commonly referred to as the morphing attack, has motivated a large and still growing body of work on Morphing Attack Detection (MAD), which is typically split into Single-image MAD (S-MAD), where only the document photograph is available, and Differential MAD (D-MAD), where a live capture of the presenting subject is additionally available for comparison~\cite{scherhag2019morphing}. The S-MAD setting is particularly relevant at the document-application stage, where the submitted photograph may need to be assessed before a trusted live reference image is available for differential comparison. As a result, a substantial research effort has gone into building S-MAD datasets and benchmarking detectors on them, most notably through the SYN-MAD 2022~\cite{huber2022syn} competition and dataset, which combines bona fide images with morphs generated by a wide range of morphing techniques, from landmark-based methods to GAN and diffusion-based approaches. Most of the existing S-MAD benchmarks, however, evaluate detectors directly on digital images, without accounting for transformations introduced by some document-application pipelines. This creates a potential domain shift in which the image presented to an S-MAD system is not the original digital image. In some passport application workflows, an applicant can submit a physical photograph that must satisfy quality requirements (e.g., ICAO 9303)~\cite{icao9303} and the issuing authority subsequently scans this photograph for digital enrolment and storage. Thus, both bona fide images and potential morphing attacks undergo a physical print-scan (P\&S) transformation before they reach an automated detector, meaning that evaluations conducted exclusively on the original digital images do not capture the acquisition domain encountered in such workflows.
The print-scan transformation is a lossy and multi-stage physical process in which printing quantizes color into a halftone pattern of ink dots, the paper substrate scatters light, and the sheet-fed scanner resamples the reflected light with its own sensor noise. Together, these effects induce an acquisition-domain shift between digital and P\&S imagery, altering the low-level texture, edge, and color statistics on which S-MAD systems may rely. As illustrated in Figure~\ref{fig:teaser}, these changes are spatially pronounced between the digital and print-scanned images, whereas the differences between the 300 and 600 DPI scans are comparatively small. Therefore, we argue that what remains missing is an open, paired and test-only benchmark on which current detectors can be compared reproducibly. Existing print-scan resources are private or sequestered~\cite{ngan2020frvtmorph, raja2021sotamd}, and recent studies evaluate few detectors or focus on training-time mitigation~\cite{ferrara2021printscan, neddo2024impact, tapia2023simulating}.
 
In this paper, we address this gap by building a print-scan counterpart dataset of SYN-MAD22~\cite{huber2022syn}. We start by physically printing every bona fide and morphed image in the dataset and re-scanning it at two resolutions, 300 and 600 DPI, following a fixed, documented acquisition protocol. We then use this new dataset to benchmark eight state-of-the-art S-MAD systems, spanning classical texture and frequency-based methods and modern CNN and transformer-based detectors. We find that state-of-the-art detectors on digital SYN-MAD22~\cite{huber2022syn} images suffer a performance collapse once the print-scan transformation is applied, and that this degradation is largely insensitive to the specific scanning resolution used (300 vs.\ 600 DPI) but strongly modulated by the morphing algorithm.
 
In summary, our main contributions are as follows:
% \begin{itemize}
%     \item We introduce MAD22-PS, a print-scanned extension of the SYN-MAD22 dataset, obtained by printing and re-scanning 5,687 bona fide and morphed images at 300 and 600 DPI, together with the full acquisition protocol to support replication and extension.
%     \item To the best of our knowledge, this is the first large-scale, controlled benchmark of a diverse panel of recent S-MAD methods on a physically print-scanned extension of SYN-MAD22, covering multiple morph-generation families and scan resolutions.
%     \item We show that print-scanning degrades S-MAD performance across eight state-of-the-art detectors, analyze which factors influence its degradation, and validate the main degradation pattern using a second scanning device.
%     \item We discuss the implications of our findings for evaluating S-MAD systems under print-scan degradation and outline directions towards print-scan–robust morphing attack detection.
% \end{itemize}

\begin{itemize}
    \item We introduce \textbf{MAD22-PS}, a paired print-scan extension of SYN-MAD22 comprising 5,687 source images (204 bona fide and 5,483 morphed), acquired using two scanners at 300 and 600 DPI. Designed as a test-only benchmark, it enables reproducible evaluation of cross-domain generalization across six morph-generation algorithms without training or fine-tuning on the benchmark.

    \item We demonstrate a substantial generalization failure across eight S-MAD detectors from diverse methodological families: average EER increases from 11.00\% on digital images to 41.85\% after print-scanning. The degradation varies dramatically across detectors, with FD-MAD increasing from 5.10\% to 77.51\% EER, showing that strong digital-domain performance does not guarantee robustness to physical acquisition.

    \item We show that degradation depends strongly on the morph-generation algorithm, whereas changing the scanning resolution from 300 to 600 DPI has little effect on average performance. We validate the overall degradation pattern using a second scanner and show that print-scanned morphs remain highly effective against four face-recognition systems, revealing that reduced detectability does not necessarily reduce attack effectiveness.
\end{itemize}
The print-scanned dataset and our evaluation code will be publicly available upon acceptance, in order to support future research on print-scan–robust morphing attack detection.
 
\section{Related Work}
\label{sec:related}
 
\vspace{0.5em}
\noindent\textbf{Face Morphing Attacks. }
Face morphing attacks were first described by Ferrara \textit{et al.}~\cite{ferrara2014magic}, who showed that landmark-based morphing of two subjects' face images could produce a single photograph that both contributors could use to pass automated and human face verification. Early morphing generation techniques relied on facial landmark detection, Delaunay triangulation, and alpha-blending~\cite{ferrara2014magic}, and were later
complemented by generative approaches based on Generative Adversarial Networks (GANs), such as MorGAN~\cite{damer2018morgan} and MIPGAN-I/II~\cite{zhang2021mipgan}, and, more recently, diffusion models~\cite{damer2023mordiff, blasingame2024greedy}, which tend to produce morphs with fewer visible blending artifacts and higher perceptual quality. This diversity of generation mechanisms is important for MAD research, since detectors that overfit to the artifacts of a single morphing technique tend to generalize poorly to unseen morphing algorithms~\cite{scherhag2019morphing}.
 
\vspace{0.5em}
\noindent\textbf{Morphing Attack Detection. }
% MAD approaches are commonly divided into S-MAD and D-MAD. D-MAD methods assume access to an additional, trusted live capture of the individual presenting the document and can therefore exploit identity-comparison cues~\cite{scherhag2019morphing}. 
% In contrast, 
S-MAD methods must determine whether a single image is a morph due to statistical artifacts left by the generation process, such as ghosting or frequency-domain inconsistencies. Proposed S-MAD approaches range from hand-crafted texture descriptors (e.g., LBP, BSIF) combined with classical classifiers, to deep CNN architectures such as MixFaceNet-MAD and Inception-MAD~\cite{damer2022privacy} and identity-disentanglement approaches~\cite{neto2022orthomad}, to more recent foundation-model-based detectors~\cite{caldeira2025madation}, self-supervised approaches~\cite{ivanovska2025selfmad, ivanovska2026selfmadpp}, and approaches that explicitly analyze the image in the frequency domain to detect blending artifacts~\cite{paulo2026fd, paulo2026srl}. The SYN-MAD 2022 competition~\cite{huber2022syn} fostered much of this progress by providing a common and large-scale benchmark consisting of bona fide images and morphs produced with a wide variety of morphing algorithms, and by ranking a large number of submitted S-MAD solutions under a standardized protocol. We build directly on this dataset in the present work.

\begin{table*}[t]
\caption{\textbf{Comparison with prior print-scan (P\&S) MAD works.} Bona fide/morph counts refer to P\&S images. n/r = not reported; -- = not studied; Seq.\ = sequestered (evaluation only via submission). P\&S train = detectors trained or fine-tuned with real or simulated P\&S data; ours is intentionally test-only, so results measure cross-domain generalization. Ours is the only open benchmark with a large panel of S-MAD models, two scanners and FR analysis.}
\label{tab:related}
\centering
\footnotesize
\setlength{\tabcolsep}{4.5pt}
\renewcommand{\arraystretch}{1.12}
% \begin{tabular}{@{}lcccccccc@{}}
\begin{tabular*}{\textwidth}{@{\extracolsep{\fill}}lcccccccc@{}}
\toprule
Related Works & Bona fide & Morphs & Morph types & Printers / scanners & MAD models & FR systems & P\&S train & Public \\
\midrule
Raghavendra \textit{et al.}~\cite{raghavendra2017transferable}
   & 352 & 431 & 1 & 1 / 2 & 1 & -- & \cmark & -- \\
NIST FRVT MORPH~\cite{ngan2020frvtmorph}
   & 2,739 & 3,604 & n/r & 1 / 1 & submitted & -- & n/r & Seq. \\
SOTAMD~\cite{raja2021sotamd}
  & 1,096 & 3,703 & multiple & multiple / multiple
  & submitted & -- & n/r & Seq. \\
Tapia \textit{et al.}\ \cite{tapia2023simulating,tapia2025printscan}
  & 984 & 3,856 & 4 & n/r / 1 & 2 & -- & \cmark & \xmark \\
Zhang \cite{zhang2024generalized}
 & 1,270 & 2,500 & 5 & 1 / 1 & 8 & -- & \cmark & \xmark \\
Patwardhan \textit{et al.}~\cite{patwardhan2025clip}
  & n/r & n/r & 5 & 2 / 1 & 4 & -- & \xmark & \xmark \\
Neddo \textit{et al.}\ \cite{neddo2024impact}
  & 4,653 & 8,142 & 3 & 1 / 1 & 1 & 3 & \cmark & \xmark \\
\midrule
\textbf{Ours (MAD22-PS)}
  & \textbf{204} & \textbf{5,483} & \textbf{6} & \textbf{1 / 2} & \textbf{8} & \textbf{4} & test-only & \textbf{\cmark} \\
\bottomrule
\end{tabular*}
\end{table*}

\vspace{0.5em}
\noindent\textbf{Print-Scan Effects on MAD. }
\label{sec:related-printscan}
Prior work has established that print-scan acquisition affects both the vulnerability of face-recognition (FR) systems to morphing attacks and the performance of MAD. Early studies showed that print-scanning harms single-image detection while morphs can still deceive FR after re-digitization~\cite{scherhag2017vulnerability,raghavendra2017transferable}. Large evaluations such as NIST FRVT MORPH~\cite{ngan2020frvtmorph}, which reported very high false-detection rates on print-scanned bona fide images for most submitted algorithms, and the SOTAMD/FVC-onGoing platform~\cite{raja2021sotamd} added print-scanned conditions for S-MAD and D-MAD. Later work largely moved to mitigation, either by training with simulated print-scan data~\cite{ferrara2021printscan,tapia2023simulating,tapia2025printscan}, which improves robustness, or by proposing ViT~\cite{dosovitskiy2020image} and CLIP-based~\cite{clip} detectors tested on digital, print-scanned and compressed variants~\cite{zhang2024generalized,patwardhan2025clip}. Table~\ref{tab:related} compares the closest works. The nearest to ours is Neddo \textit{et al.}~\cite{neddo2024impact}, who print-scan three datasets and analyze mixed digital/print-scan pairings for three FR systems and one S-MAD network that they train themselves. We instead keep the subjects, morph pairs and protocol of the established SYN-MAD22 benchmark~\cite{huber2022syn} fixed, and evaluate eight off-the-shelf detectors from different families on its paired print-scanned counterpart, using two scanners, two resolutions, six morph generators (including diffusion morphs) and four FR systems. To the best of our knowledge, MAD22-PS is also the only one of these resources that is publicly released. The NIST and SOTAMD print-scan data are sequestered and accessible only through algorithm submission, which prevents score-level analysis and reproducible comparison. By design, MAD22-PS is a test-only benchmark for cross-domain generalization, so print-scan-aware training and mixed pairings are complementary and out of its scope.

\vspace{0.5em}
\noindent\textbf{Robustness of Biometric Systems to Real-World Degradations. }
More broadly, a parallel and growing line of work has shown that biometric recognition and
detection systems, when evaluated only on curated and clean benchmark data, overestimate the accuracy that can be expected under real-world acquisition conditions. In face recognition, occlusions such as masks, sunglasses, and scarves have been shown to disproportionately harm accuracy and, more importantly, to do so unequally across demographic groups, exacerbating existing fairness gaps~\cite{mamede2024fairness}. Similarly, in~\cite{tian2024impact} it was shown that the presence of dark sunglasses in a probe image degrades one-to-many facial identification accuracy by an amount comparable to severe Gaussian blur or a substantial reduction in image resolution, and that this degradation compounds when sunglasses are combined with other quality factors. Our work is directly motivated by this broader body of work on the fragility of biometric systems to distribution shifts and extends the specific print-scan MAD literature along the axes of dataset scale, resolution comparison, detector recency, and joint MAD/FR-vulnerability analysis.

\section{Print-Scanned SYN-MAD22 Dataset}
\label{sec:dataset}
MAD22-PS is a physical printed and scanned extension of the SYN-MAD22 benchmark~\cite{huber2022syn}, designed to preserve its underlying subjects, bona fide samples, and morph-generation methods while introducing controlled physical acquisition conditions. Our dataset contains the original digital images together with corresponding print-scanned versions acquired at 300 and 600 DPI and satisfies ICAO passport-photo compliance constraints. From SYN-MAD22, MAD22-PS contains 204 bona fide images and 5,483 morphed images generated using OpenCV, FaceMorpher, WebMorph, MIPGAN-I, MIPGAN-II~\cite{zhang2021mipgan} and MorDIFF~\cite{damer2023mordiff}, comprising a total of 5,687 images. 
% For the vulnerability analysis in Section~\ref{sec:results}, we additionally evaluate digital MorDIFF morphs~\cite{damer2023mordiff} although these are not currently included in the physically print-scanned portion of MAD22-PS.

% \vspace{-.5em}
\subsection{Print-Scan Acquisition Protocol}
\begin{figure*}[t]
    \centering
    \includegraphics[width=0.95\linewidth]{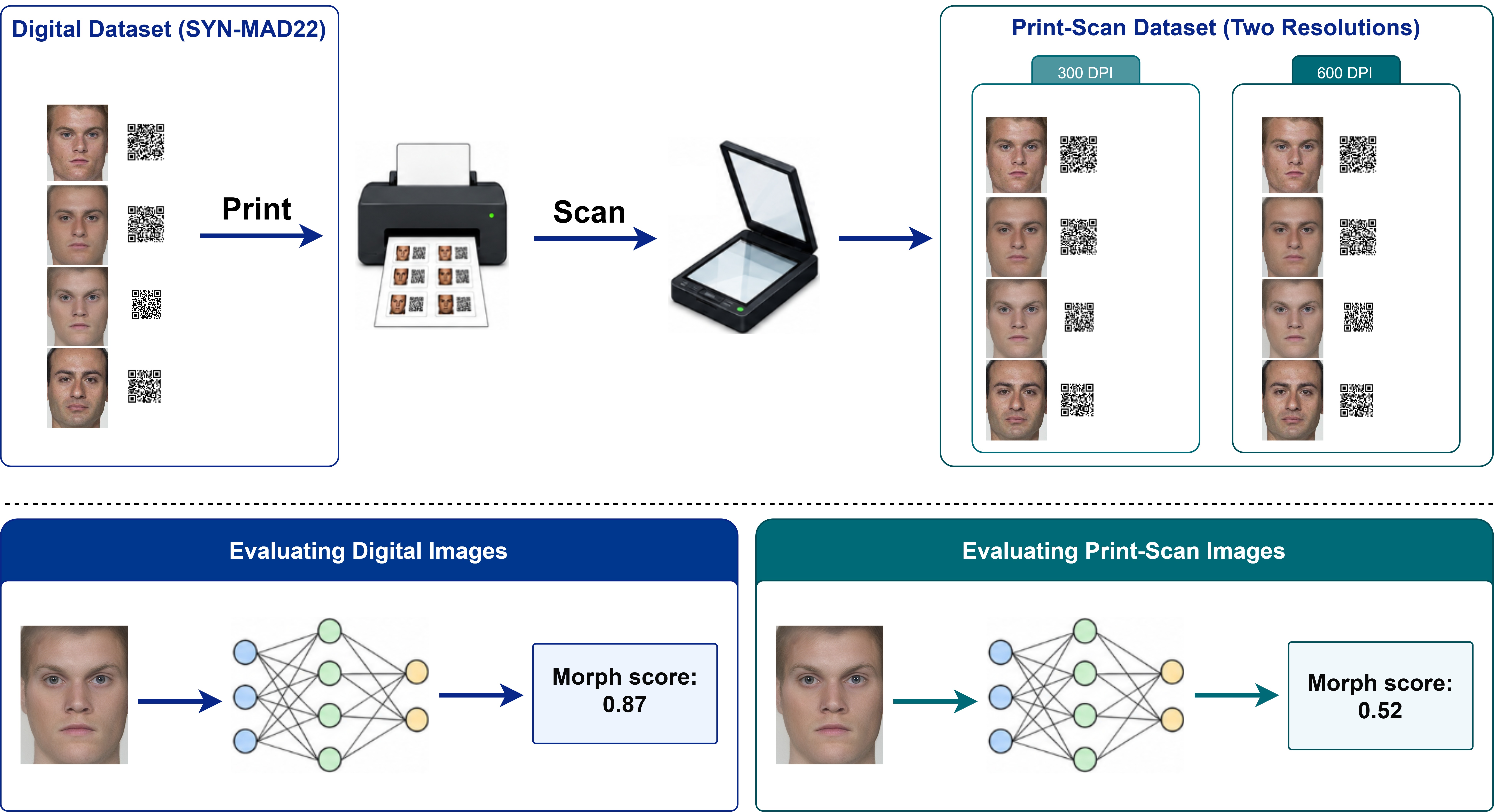}
    \caption{\textbf{Print-scan acquisition and evaluation pipeline.} Digital SYN-MAD22~\cite{huber2022syn} images (bona fide and morphed) are arranged in groups of four on A4 sheets and printed on a laser-jet printer. Each image is accompanied by a QR code containing the metadata required to associate the scanned image with its original digital sample. The sheets are re-scanned at 300 and 600 DPI using a multifunction scanner and subsequently automatically detected, cropped, and geometrically aligned to the original ICAO-compliant crop. The resulting images are used to evaluate eight S-MAD detectors, and their results are compared to the digital images as a reference condition.}
    \label{fig:pipeline}
% \vspace{-1em}
    
\end{figure*}

To construct our print-scan counterpart dataset, every image in the selected
SYN-MAD22 dataset was physically printed and subsequently re-scanned,
following the pipeline illustrated in Figure~\ref{fig:pipeline}.
 
\vspace{0.5em}
\noindent\textbf{Printing.} Four photographs were arranged on each A4 sheet accompanied by a QR code encoding its source-image filename, its file extension, and morph-generation algorithm. The QR codes were used only to associate each recovered photograph with the corresponding digital image and morphing method; thus, they were not included in the image crops used for evaluation. The sheets were printed using a Xerox PrimeLink C9065 on 180\,g/m$^2$ glossy photo paper, at the recommended ICAO passport-photo print size of 70$\times$100\,mm, using the printer's default color management profile.

\vspace{0.5em}
\noindent\textbf{Scanning.} The printed A4 sheets were re-digitized using a Konica Minolta bizhub C250i sheet-fed/multifunction scanner, at two resolutions commonly used in administrative and passport-office digitization workflows: 300 DPI and 600 DPI. All scans were acquired in 24-bit RGB color and stored as PDF 1.4 without lossy image compression. To assess whether the observed print--scan effects are specific to a single scanning device, we additionally repeated the scanning stage at both 300 and 600 DPI using a Konica Minolta bizhub C450i. The same post-processing and evaluation protocols were applied to both scanner acquisitions. Unless otherwise stated, the detailed analyses in the main paper use the C250i acquisition, while the C450i acquisition is used as a cross-scanner robustness check.

\vspace{0.5em}
\noindent\textbf{Post-processing.} Each scanned A4 sheet was automatically processed to recover the individual photographs and their associated QR codes. The extracted print-scanned image was saved, using the same filename and extension as its corresponding digital image, and corrected for any residual rotation introduced during scanning. For the S-MAD evaluation, the cropping protocol described in~\cite{damer2022privacy} was applied identically to the digital and print-scanned versions of each sample. No additional denoising, sharpening, or contrast enhancement was applied, in order to preserve the photometric characteristics introduced by
the print--scan process.

\subsection{Resulting Dataset Statistics}

MAD22-PS contains 5,687 source samples represented under five acquisition conditions: the original digital condition and four print--scan conditions obtained using two scanners at 300 and 600 DPI.
Each acquisition condition therefore contains the same 204 bona fide and 5,483 morphed samples, yielding 28,435 image instances in total. Table~\ref{tab:dataset-stats} summarizes the dataset composition.

\begin{table}[t]
\centering
\caption{\textbf{Composition of MAD22-PS.}
The four physical print--scan conditions correspond to two scanners
(C250i and C450i) at 300 and 600 DPI. Morphs span six generation
algorithms: OpenCV, FaceMorpher, WebMorph,
MIPGAN-I~\cite{zhang2021mipgan},
MIPGAN-II~\cite{zhang2021mipgan} and
MorDIFF~\cite{damer2023mordiff}.}
\label{tab:dataset-stats}

\resizebox{\linewidth}{!}{%
\begin{tabular}{lcccc}
\toprule
\textbf{Subset}
& \textbf{Conditions}
& \textbf{Bona fide }
& \textbf{Morphs}
& \textbf{Image instances} \\
\midrule
Digital
& 1
& 204
& 5,483
& 5,687 \\

Print--scan
& 4
& 204
& 5,483
& 22,748 \\
\midrule
\textbf{Total}
& \textbf{5}
& --
& --
& \textbf{28,435} \\
\bottomrule
\end{tabular}%
}

% \vspace{-1em}
\end{table}

% We will publicly release the print-scanned images, the acquisition
% protocol described above, and all evaluation code upon publication,
% to support replication and further research on print-scan--robust MAD.
\subsection{Availability and Licensing}
MAD22-PS is derived from the MAD22 benchmark of SYN-MAD~2022~\cite{huber2022syn}, whose bona fide images come from the Face Research Lab London (FRLL) set~\cite{frll}. FRLL is released under the CC~BY~4.0 license, and its participants consented to their images being used in research. Our acquisition only re-digitizes existing images (print and scan) and involves no new human subjects, so no additional identities or personal data are introduced. Therefore, access to the digital morph images follows the MAD22 request procedure, and we distribute only the print-scanned version respecting the applicable source-dataset terms. The print-scanned dataset and our evaluation code will be publicly available upon acceptance, in order to support future research on print-scan–robust morphing attack detection.

\section{Experimental Design}
\label{sec:method}
 
\subsection{Research Questions}
Our evaluation is designed to answer three questions: \textbf{(RQ1)} How
much does print-scanning degrade the accuracy of state-of-the-art S-MAD
detectors relative to their digital-domain performance? \textbf{(RQ2)} Does
this degradation depend on the scanning resolution (300 vs.\ 600 DPI)?
\textbf{(RQ3)} Does the degradation depend on the morphing algorithm used
to generate the attack image, i.e., are morphs produced by some generation
techniques inherently more (or less) robust to print-scanning than others,
and does this degradation in detection accuracy translate into a corresponding drop in attack effectiveness against face recognition systems?
 
\subsection{S-MAD Models}
We evaluate a panel of eight S-MAD models, selected to be representative of the main families of approaches in the current literature. We use MixFaceNet-MAD~\cite{damer2022privacy} and Inception-MAD~\cite{damer2022privacy}, which are two lightweight CNN-based detectors, OrthoMAD~\cite{neto2022orthomad}, a ResNet-18-based detector that learns to disentangle the two contributing identities into orthogonal latent vectors, PCA-MAD, a classical baseline that performs morph detection on a PCA-reduced texture representation of the face image, MADation (ViT-B) and MADation (ViT-L)~\cite{caldeira2025madation}, foundation-model-based detectors that adapt a CLIP~\cite{clip} vision encoder with LoRA~\cite{hu2022lora} weights, FD-MAD~\cite{paulo2026fd}, a lightweight, interpretable detector based on residual Fourier-domain features extracted from different facial regions and SelfMAD++~\cite{ivanovska2026selfmadpp}, a self-supervised foundation-model-based detector that learns to detect generic morphing artifacts simulated during training. Where possible, we use publicly available pre-trained weights to avoid introducing confounds from re-training. Furthermore, all detectors are trained exclusively on digital data (SMDD), so that none of them has ever observed a print-scanned image during training. This is a deliberate design choice since our goal is to measure the out-of-domain performance of existing digitally trained S-MAD models when evaluated without adaptation on MAD22-PS, not the accuracy that could be recovered by retraining on print-scanned data (which we discuss as a mitigation strategy in Section~\ref{sec:discussion}). Training on MAD22-PS would turn this into an in-domain evaluation and make results across methods incomparable. We therefore release the set with a fixed protocol (no training, fine-tuning or threshold selection on MAD22-PS) and recommend that print-scan-robust detectors be trained on separate print-scanned or simulated data~\cite{tapia2023simulating,tapia2025printscan,ferrara2021printscan} and then evaluated on MAD22-PS.

\subsection{Evaluation Protocol and Metrics}
\label{sec:metrics}
Following standard practice in the MAD literature and the ISO/IEC 30107-3~\cite{iso30107} presentation attack detection standard, we report Equal Error Rate (EER), which corresponds to the operating point at which APCER equals BPCER. Additionally, we report the Mated Morph Presentation Match Rate (MMPMR), which measures the proportion of mated morph comparisons that are accepted by a face recognition system at a specified decision threshold. We report MMPMR at false match rate (FMR) operating points of $1\%$ and $0.1\%$ in Table~\ref{tab:morphing_ps}. The corresponding model-specific thresholds are estimated on LFW~\cite{lfw}, following the SYN-MAD 2022 protocol~\cite{huber2022syn}, and are applied unchanged to all acquisition conditions. For each detector, morphing algorithm, and acquisition condition, we obtain EER directly from the corresponding evaluation scores by sweeping the decision threshold and identifying the operating point at which APCER and BPCER are equal. We use EER as a threshold-independent summary of the separability between bona fide and morphed score distributions, instead of using a fixed operational decision threshold.

% Maybe remove it? 
% \subsection{Experimental Conditions}
% For every detector, we evaluate three acquisition conditions: (1) Digital, the SYN-MAD22~\cite{huber2022syn} images (reference condition); (2) Print-Scan 300 (PS300); and (3) Print-Scan 600 (PS600). For the face recognition vulnerability analysis (Table~\ref{tab:morphing_ps}), we additionally distinguish between image-level morphs (OpenCV, FaceMorpher, WebMorph), which blend pixel intensities directly, and representation-level morphs (MIPGAN-I~\cite{zhang2021mipgan}, MIPGAN-II~\cite{zhang2021mipgan}, MorDIFF~\cite{damer2023mordiff}), which blend the two identities in a GAN or diffusion latent space before decoding, since these two families of attacks are known to leave visually and statistically distinct traces.

\section{Results}
\label{sec:results}
In order to quantify the impact of print--scan acquisition, we organize the results around the three research questions. RQ1 examines the overall degradation of S-MAD accuracy relative to the original digital condition. RQ2 investigates whether this degradation is affected by the scanning resolution, comparing 300 and 600 DPI. RQ3 analyzes the importance of the morph-generation algorithm and examines whether reduced morph detectability is accompanied by a corresponding reduction in the effectiveness of the attacks against downstream face recognition systems. We first report the EER results for the eight S-MAD models and subsequently analyze the MMPMR results for four face recognition models. Unless stated otherwise, the digital condition denotes the original SYN-MAD22~\cite{huber2022syn} images, whereas PS300 and PS600 denote the corresponding images after printing and re-scanning at 300 and 600 DPI, respectively.

\begin{table*}[t]
\centering
\small
\setlength{\tabcolsep}{4pt}
\caption{\textbf{EER (\%, $\downarrow$) of eight S-MAD models under
digital and print--scan acquisition on SYN-MAD22~\cite{huber2022syn} using the Konica Minolta bizhub C250i.}
PS600 and PS300 denote print--scan acquisition at 600 and 300 dpi,
respectively. Morphing algorithms are grouped according to whether
morph generation operates at the image or representation level.}
\label{tab:mad_printscan}
\begin{tabular*}{\textwidth}{
@{\extracolsep{\fill}}lcccccccc@{}
}
% \toprule
\cmidrule[\heavyrulewidth]{1-9} 
\multirow{2}{*}{
\begin{tabular}{c}
S-MAD Detector
\end{tabular}
}
& \multirow{2}{*}{Acquisition}
& \multicolumn{3}{c}{Image level morphing}
& \multicolumn{3}{c}{Representation level morphing}
& \multirow{2}{*}{Average} \\
\cmidrule(lr){3-5}
\cmidrule(lr){6-8}
&
& OpenCV
& FaceMorpher
& WebMorph
& MIPGAN-I~\cite{zhang2021mipgan}
& MIPGAN-II~\cite{zhang2021mipgan}
& MorDIFF~\cite{damer2023mordiff}
& \\
\cmidrule{1-9}

\multirow{3}{*}{MixFaceNet-MAD~\cite{damer2022privacy}}
& Digital        & 8.33  & 4.60  & 18.20 & 16.70  & 20.62  & 8.50  & 12.83 \\
& \textbf{PS600} & 15.04 & 6.82  & 26.96 & 43.14  & 44.04  & 84.97 & 36.83 \\
& \textbf{PS300} & 15.20 & 6.86  & 28.92 & 28.43  & 31.37  & 76.49 & 31.21 \\
\cmidrule{1-9}

\multirow{3}{*}{Inception-MAD~\cite{damer2022privacy}}
& Digital        & 7.52  & 0.00  & 18.00 & 10.90  & 16.22  & 5.30  & 9.66 \\
& \textbf{PS600} & 9.51  & 2.60  & 30.76 & 47.84  & 48.19  & 48.61 & 31.25 \\
& \textbf{PS300} & 11.55 & 2.60  & 42.51 & 49.45  & 49.79  & 48.03 & 33.99 \\
\cmidrule{1-9}

\multirow{3}{*}{OrthoMAD~\cite{neto2022orthomad}}
& Digital        & 4.07  & 0.10  & 13.80 & 11.20  & 12.21  & 10.80 & 8.70 \\
& \textbf{PS600} & 10.26 & 5.90  & 26.40 & 33.90  & 30.73  & 89.50 & 32.78 \\
& \textbf{PS300} & 10.67 & 4.20  & 28.00 & 27.60  & 24.02  & 87.80 & 30.38 \\
\cmidrule{1-9}

\multirow{3}{*}{PCA-MAD~\cite{dargaud2023principal}}
& Digital        & 28.50 & 0.49  & 34.82 & 49.90  & 47.13  & 0.10  & 26.82 \\
& \textbf{PS600} & 38.11 & 62.51 & 45.10 & 71.54  & 71.46  & 71.48 & 60.03 \\
& \textbf{PS300} & 35.78 & 60.29 & 42.80 & 81.06  & 80.17  & 81.49 & 63.60 \\
\cmidrule{1-9}

\multirow{3}{*}{MADation (ViT-B)~\cite{caldeira2025madation}}
& Digital        & 3.85  & 0.00  & 10.80 & 33.37  & 22.21  & 1.10  & 11.89 \\
& \textbf{PS600} & 34.75 & 5.80  & 51.00 & 88.90  & 81.88  & 59.00 & 53.56 \\
& \textbf{PS300} & 34.45 & 7.80  & 52.20 & 90.00  & 83.98  & 68.50 & 56.16 \\
\cmidrule{1-9}

\multirow{3}{*}{MADation (ViT-L)~\cite{caldeira2025madation}}
& Digital        & 19.26 & 0.40  & 2.23  & 20.32  & 9.06   & 20.40 & 11.95 \\
& \textbf{PS600} & 38.21 & 23.50 & 51.80 & 34.00  & 34.73  & 31.30 & 35.59 \\
& \textbf{PS300} & 39.84 & 22.00 & 52.80 & 31.90  & 33.13  & 27.00 & 34.45 \\
\cmidrule{1-9}

\multirow{3}{*}{FD-MAD~\cite{paulo2026fd}}
& Digital        & 2.54  & 0.00  & 18.40 & 2.45   & 7.21   & 0.00   & 5.10 \\
& \textbf{PS600} & 27.45 & 99.60 & 38.00 & 100.00 & 100.00 & 100.00 & 77.51 \\
& \textbf{PS300} & 26.47 & 89.20 & 37.80 & 100.00 & 100.00 & 100.00 & 75.58 \\
\cmidrule{1-9}

\multirow{3}{*}{SelfMAD++~\cite{ivanovska2026selfmadpp}}
& Digital        & 0.00  & 0.00  & 6.37  & 0.00   & 0.00   & 0.00 & 1.06 \\
& \textbf{PS600} & 6.86  & 0.49  & 26.47 & 7.35   & 1.47   & 0.98 & 7.27 \\
& \textbf{PS300} & 8.82  & 0.00  & 27.45 & 10.78  & 2.45   & 0.98 & 8.41 \\
\cmidrule{1-9}

\multirow{3}{*}{Average}
& Digital        & 9.26  & 0.70  & 15.33 & 18.11  & 16.83  & 5.78  & 11.00 \\
& \textbf{PS600} & 22.52 & 25.90 & 37.06 & 53.33  & 51.56  & 60.73 & 41.85 \\
& \textbf{PS300} & 22.85 & 24.12 & 39.06 & 52.40  & 50.61  & 61.29 & 41.72 \\
% \bottomrule
\cmidrule[\heavyrulewidth]{1-9} 
\end{tabular*}
\end{table*}

\begin{table}[t]
\centering
\small
\setlength{\tabcolsep}{6pt}
\caption{\textbf{Cross-scanner comparison of S-MAD performance.}
Average EER (\%, $\downarrow$) across eight detectors and six
morphing algorithms. Results are reported for the original digital
images and for print--scan acquisition using two scanners at
300 and 600 dpi.}
\label{tab:cross_scanner}
\begin{tabular}{lccccc}
\toprule
& Digital
& \multicolumn{2}{c}{C250i}
& \multicolumn{2}{c}{C450i} \\
\cmidrule(lr){3-4}
\cmidrule(lr){5-6}
& & PS300 & PS600 & PS300 & PS600 \\
\midrule
Average EER
& 11.00
& 41.72
& 41.85
& 39.78
& 38.96 \\
\bottomrule
\end{tabular}
% \vspace{-1em}
\end{table}

\begin{table*}[t]
\centering
\footnotesize
\setlength{\tabcolsep}{4pt}
\renewcommand{\arraystretch}{1.05}
\caption{\textbf{Morphing vulnerability of four face recognition (FR)
systems under digital and print--scan acquisition using the Konica Minolta bizhub C250i.}
Each cell reports MMPMR100/MMPMR1000 ($\uparrow$; higher means the
morph is a more successful attack). PS600 and PS300 indicate
print--scan acquisition at 600 and 300 dpi, respectively.
Morphing algorithms are grouped according to whether morph generation
operates at the image or representation level.}
\label{tab:morphing_ps}

% \begin{tabular}{lccccccc}
\begin{tabular*}{\textwidth}{
@{\extracolsep{\fill}}lccccccc@{}
}
\toprule
\multirow{2}{*}{FR system}
& \multirow{2}{*}{Acquisition}
& \multicolumn{3}{c}{Image level morphing}
& \multicolumn{3}{c}{Representation level morphing} \\
\cmidrule(lr){3-5}
\cmidrule(lr){6-8}
&
& OpenCV
& FaceMorpher
& WebMorph
& MIPGAN-I~\cite{zhang2021mipgan}
& MIPGAN-II~\cite{zhang2021mipgan}
& MorDIFF~\cite{damer2023mordiff} \\
\midrule

\multirow{3}{*}{CurricularFace~\cite{huang2020curricularface}}
& Digital
& 0.996/0.986
& 0.962/0.913
& 0.990/0.986
& 0.962/0.890
& 0.953/0.832
& 0.995/0.986 \\
& PS600
& 0.996/0.986
& 0.974/0.934
& 0.992/0.988
& 0.955/0.862
& 0.935/0.812
& 0.993/0.971 \\
& PS300
& 1.000/0.987
& 0.972/0.930
& 0.992/0.986
& 0.960/0.874
& 0.942/0.826
& 0.993/0.969 \\
\cmidrule{1-8}

\multirow{3}{*}{ElasticFace~\cite{boutros2022elasticface}}
& Digital
& 0.997/0.980
& 0.962/0.913
& 0.990/0.986
& 0.980/0.845
& 0.953/0.778
& 0.990/0.948 \\
& PS600
& 0.999/0.988
& 0.965/0.939
& 0.990/0.986
& 0.965/0.896
& 0.961/0.888
& 0.992/0.967 \\
& PS300
& 0.997/0.989
& 0.957/0.930
& 0.990/0.982
& 0.967/0.876
& 0.960/0.831
& 0.992/0.974 \\
\cmidrule{1-8}

\multirow{3}{*}{MagFace~\cite{meng2021magface}}
& Digital
& 0.995/0.966
& 0.953/0.911
& 0.988/0.984
& 0.951/0.835
& 0.934/0.766
& 0.986/0.935 \\
& PS600
& 0.999/0.995
& 0.979/0.955
& 0.990/0.986
& 0.984/0.943
& 0.974/0.935
& 0.993/0.961 \\
& PS300
& 0.998/0.983
& 0.963/0.929
& 0.988/0.982
& 0.966/0.881
& 0.958/0.840
& 0.989/0.963 \\
\cmidrule{1-8}

\multirow{3}{*}{ArcFace~\cite{deng2019arcface}}
& Digital
& 0.993/0.956
& 0.951/0.898
& 0.988/0.984
& 0.945/0.770
& 0.930/0.708
& 0.987/0.923 \\
& PS600
& 1.000/0.988
& 0.976/0.939
& 0.990/0.988
& 0.970/0.902
& 0.964/0.892
& 0.993/0.961 \\
& PS300
& 0.997/0.985
& 0.966/0.929
& 0.988/0.986
& 0.969/0.881
& 0.956/0.832
& 0.995/0.966 \\

\bottomrule
\end{tabular*}
\end{table*}
 
\subsection{RQ1: Overall Impact of Print-Scanning}
Table~\ref{tab:mad_printscan} reports the EER of eight S-MAD models in the digital domain and after print-scan acquisition at 600 and 300 DPI, respectively. Averaged across all eight detectors and six morphing algorithms, EER rises from 11.00\% in the digital domain to 41.85\% at 600 DPI and 41.72\% at 300 DPI, which represents more than a threefold relative increase (+280\% and +279\%, respectively) and indicates that digital-domain performance does not transfer reliably to print-scanned images. Every single detector in our analysis exhibits an increase in EER under print-scanning, without exception. The smallest degradation is observed for SelfMAD++~\cite{ivanovska2026selfmadpp}, whose EER increases from 1.06\% to 7.27\% (PS600) / 8.41\% (PS300), which represents a six to eightfold relative increase. The largest, and most concerning, degradation is observed for FD-MAD~\cite{paulo2026fd}, whose EER increases from 5.10\%, the second-best digital result in our analysis, to 77.51\% (PS600) / 75.58\% (PS300). These results indicate a domain shift and a catastrophic failure when the digitally trained
detector is tested on print-scanned images. The sensitivity of FD-MAD is consistent with its reliance
on residual Fourier-domain features extracted from different facial regions, and the printing halftone process, together with the scanner's own resampling and anti-aliasing filtering, constitutes a strong and highly structured perturbation of exactly the mid-to-high spatial frequencies that such detectors exploit. This confirms our hypothesis that print-scanning degrades the accuracy of state-of-the-art S-MAD models and shows that strong digital-domain performance is not just imperfect, but potentially misleading, as a predictor of print-scan robustness.
 
\subsection{RQ2: Effect of Scanning Resolution}
Comparing the PS600 and PS300 columns of Table~\ref{tab:mad_printscan} shows that scanning resolution has, at best, a second-order effect on S-MAD accuracy: the panel-averaged EER differs by only 0.13 percentage points between 600 DPI (41.85\%) and 300 DPI (41.72\%), and four out of eight detectors are, in fact, marginally more accurate at 300 DPI than at 600 DPI (e.g., MixFaceNet-MAD: 36.83\% at PS600 vs. 31.21\% at PS300). This indicates that the dominant source of degradation is not the resolution at which the scanner samples the printed photograph, but rather the printing or scanning step itself, which contains factors such as halftoning, dot gain, and paper-substrate light scattering, together with lower-level color and contrast transformations that are common to both DPI settings. Within the evaluated acquisition setup, increasing the scanning resolution from 300 to 600 DPI does not meaningfully improve average S-MAD performance. This suggests that scanning resolution alone is insufficient to address the observed print-scan domain shift.

\vspace{0.5em}
\noindent\textbf{Cross-scanner robustness. } To examine whether the observed degradation is specific to the scanner used in the primary acquisition, we repeated the evaluation using a second scanning device, the Konica Minolta bizhub C450i. As summarized in Table~\ref{tab:cross_scanner}, the increase in average EER relative to the digital condition is observed with both scanners at both resolutions. Although the precise EER values vary between devices, the dominant performance loss is associated with print--scan acquisition rather than with a particular scanner or DPI setting. The complete per-detector and per-morphing-algorithm results for the C450i are reported in the supplementary material (Table~1).
 
\subsection{RQ3: Effect of the Morphing Algorithm}
Table~\ref{tab:mad_printscan} additionally breaks results down by morphing generation algorithm, averaged across detectors, and we observe that two patterns emerge. First, the digital-domain difficulty of each algorithm is not preserved under print-scanning; that is, morphs produced by FaceMorpher, which are almost perfectly detected in the digital domain (0.70\% EER), become substantially harder to detect once printed and scanned (25.90\% at PS600 / 24.12\% at PS300). This represents more than a 30-fold relative increase. Second, representation-level morphs exhibit a higher degradation under print-scanning. Most notably, MorDIFF~\cite{damer2023mordiff} increases from 5.78\% EER in the digital domain to 60.73\% at PS600 and 61.29\% at PS300, representing more than a tenfold relative increase and the highest average print-scan EER among the evaluated morphing algorithms. MIPGAN-I and MIPGAN-II~\cite{zhang2021mipgan} similarly exceed 50\% EER after print-scanning, which indicates that the digital-domain detectability of representation-level morphs does not necessarily transfer to the physical acquisition domain. The nearly identical MorDIFF results at 300 and 600 DPI further support the RQ2 observation that scanning resolution has a comparatively limited effect. For the GAN-based morphs, we attribute this effect to the domain shift and to the fact that GAN-generated morphs already minimize the low-level pixel-domain blending artifacts (ghosting, double edges) that some detectors exploit. Print-scanning additionally modifies residual texture and frequency information through halftoning, resampling, and scanner processing. In contrast, image level/landmark-based morphs remain comparatively more detectable after print-scanning, but still suffer large relative degradations across every detector in our analysis. % Consider removing the next sentence about MorDIFF if I manage to obtain this data.
% It is worth mentioning that we were unable to include print-scanned MorDIFF~\cite{damer2023mordiff} samples in the current version of the dataset; extending the print-scan acquisition to diffusion-based morphs is an immediate direction for future work.

\subsection{Attack Success vs. Detection Failure}
Beyond MAD accuracy, Table~\ref{tab:morphing_ps} reports the MMPMR, the standard metric for the effectiveness of a morph as an attack against a face recognition system (FRs). This metric was reported for four modern FR models, namely CurricularFace~\cite{huang2020curricularface}, ElasticFace~\cite{boutros2022elasticface}, MagFace~\cite{meng2021magface}, and ArcFace~\cite{deng2019arcface}. For each FR model, the similarity-score thresholds corresponding to the FMR operating points of $1\%$ and $0.1\%$ are estimated on the LFW benchmark~\cite{lfw}, following the evaluation protocol of SYN-MAD 2022~\cite{huber2022syn}. 

Overall, morphing attacks remain highly effective and sometimes even obtain higher MMPMR values (e.g., FaceMorpher with MMPMR100 of 0.962 at Digital vs. 0.972 at PS300) after print-scanning. A possible explanation is that the print--scan process could act as a smoothing and resampling transformation that attenuates fine and high-frequency morphing artifacts while preserving much of the low and mid-frequency facial structure associated with identity. Since S-MAD methods often depend on these fine artifacts, their accuracy decreases. In contrast, face recognition models can remain sensitive to the coarser identity information shared by the morph and its contributors. The suppression of unnatural local details may even make the morph representation more compatible with both identities, leading to slightly higher MMPMR values. Hence, print-scanning can make morphs harder to detect without making them less effective against the downstream face recognition system. Additionally, we report the FR-vulnerability evaluation using the C450i scanner in the supplementary material (Table~2), and we observe that morphs remain highly effective against the evaluated FR systems after print-scanning. 
 
%%%%%%%%%%%%%%%%%%%%%%%%%%%%%%%%%%%%%%%%%%%%%%%%%%%%%%%%%%%%%%%%%%%%%%%%%%%%

\section{Discussion}
\label{sec:discussion}
Our results indicate that the strong performance reported for S-MAD methods on digital benchmarks such as SYN-MAD22 does not transfer to the MAD22-PS test conditions. Based on this, we identify three implications for S-MAD evaluation. First, performance on digital benchmarks should not be assumed to transfer to print-scanned images. Evaluations should therefore report performance on print-scanned test data alongside digital test performance. For instance, NIST's own FRVT MORPH evaluation~\cite{ngan2020frvtmorph} reached a similar conclusion using a different dataset and detector pool, and our results show that this concern remains unresolved for a substantially more recent generation of S-MAD algorithms. Second, our per-resolution results (RQ2) show that increasing the scanning DPI is, on its own, an ineffective mitigation since this degradation is driven predominantly by the printing or scanning step, not by scanner sampling fidelity. This conclusion is further supported by the cross-scanner experiment, which shows that although the precise degradation varies between the C250i and C450i, both devices exhibit a loss of S-MAD accuracy relative to the digital condition, which indicates that the observed domain shift is not specific to a single scanner. Third, print-scanning degrades S-MAD performance while preserving the effectiveness of the morphing attacks against downstream FR systems. Hence, under the MAD22-PS test conditions, the attacks become harder to detect without becoming less effective against face recognition systems. We also note that the direction of the print-scan-induced error is not consistent across the literature; for instance, NIST~\cite{ngan2020frvtmorph} observed predominantly elevated false-detection rates on bona fide images, whereas our worst-affected detector, FD-MAD~\cite{paulo2026fd}, instead misses morphs at an elevated rate. This suggests that print-scan robustness cannot be assumed to fail "safely" (i.e., towards flagging more images for manual review) and must be empirically verified per detector instead of inferring from the failure mode of a single prior evaluation. Based on these findings, we point to two complementary directions for closing this gap. The first is data-centric, that is, augmenting S-MAD training sets with print-scanned (or simulated print-scan) samples, in the same spirit as in~\cite{tapia2023simulating, tapia2025printscan}, where synthetic print-scan texture transfer is used for MAD training, and as prior work that recovered part of the accuracy lost to sunglasses occlusion in face recognition by augmenting the gallery/training set with the degradation itself~\cite{tian2024impact}. Furthermore, our finding that SelfMAD++~\cite{ivanovska2026selfmadpp}, whose training procedure already relies on simulating generic morphing artifacts instead of memorizing algorithm-specific traces, degrades less than the other seven detectors (RQ1) is encouraging initial evidence that artifact-simulation-based training may transfer some robustness to the print-scan domain even without ever observing a real print-scanned image, and, combined with the results presented in~\cite{tapia2023simulating}, suggests that explicitly incorporating a print-scan simulation module into the artifact-simulation pipeline could close much of the remaining gap. The second direction is model-centric, that is, designing detectors or explicit print-scan simulation modules that are invariant to halftoning and re-sampling noise, while remaining sensitive to genuine morphing artifacts. This is a harder objective, since, as we show in Section~\ref{sec:results}, the print-scan process appears to remove or mask exactly the kind of signals that S-MAD systems rely on most heavily.

\vspace{0.5em}
\noindent\textbf{Limitations.} Although our study is based on two different scanners with two DPI settings, it uses a single printer (Xerox PrimeLink C9065). Therefore, different hardware or paper stocks could produce different degradation profiles, and due to this, we encourage future work to broaden this hardware coverage. Furthermore, since the evaluation protocol deliberately excludes training or fine-tuning on MAD22-PS, our results quantify the generalization of
digitally trained detectors instead of the performance achievable
through print-scan-aware adaptation. Investigating such adaptation
strategies remains an important direction for future work.
 
%%%%%%%%%%%%%%%%%%%%%%%%%%%%%%%%%%%%%%%%%%%%%%%%%%%%%%%%%%%%%%%%%%%%%%%%%%%%
\section{Conclusion}
\label{sec:conclusion}
We introduced MAD22-PS, a print-scanned extension of SYN-MAD22~\cite{huber2022syn} at 300 and 600 DPI, and evaluated eight state-of-the-art S-MAD systems. Average EER increased from 11.00\% on digital images to approximately 42.00\% after print-scanning, with especially severe failures for the frequency-domain method. The degradation was largely insensitive to scanning resolution but varied more across morphing algorithms. At the same time, morphs remained highly effective against the evaluated face recognition systems, showing that print-scanning can make attacks harder to detect without reducing their effectiveness. These results demonstrate that digital-only benchmarks can substantially overestimate operational robustness. We release MAD22-PS and the evaluation protocol to support the development of print-scan–robust S-MAD systems.

%%%%%%%%%%%%%%%%%%%%%%%%%%%%%%%%%%%%%%%%%%%%%%%%%%%%%%%%%%%%%%%%%%%%%%%%%%%%

{\small
\bibliographystyle{ieee}
\bibliography{egbib}

@String(CVPR= {IEEE Conf. Comput. Vis. Pattern Recog.})

@String(ICLR = {Int. Conf. Learn. Represent.})

@String(CVPRW= {IEEE Conf. Comput. Vis. Pattern Recog. Worksh.})

@String(CVPR  = {CVPR})

@String(ICLR  = {ICLR})

@String(CVPRW= {CVPRW})

@inproceedings{ferrara2014magic,
  title={The magic passport},
  author={Ferrara, Matteo and Franco, Annalisa and Maltoni, Davide},
  booktitle={IEEE International Joint Conference on Biometrics (IJCB)},
  pages={1--7},
  year={2014},
}

@article{scherhag2019morphing,
  title={Face recognition systems under morphing attacks: A survey},
  author={Scherhag, Ulrich and Rathgeb, Christian and Merkle, Johannes and Breithaupt, Ralph and Busch, Christoph},
  journal={IEEE Access},
  volume={7},
  pages={23012--23026},
  year={2019},
  publisher={IEEE}
}

@inproceedings{huber2022syn,
  title={SYN-MAD 2022: Competition on face morphing attack detection based on privacy-aware synthetic training data},
  author={Huber, Marco and Boutros, Fadi and Kirchbuchner, Florian and Damer, Naser and others},
  booktitle={IEEE International Joint Conference on Biometrics (IJCB)},
  pages={1--10},
  year={2022},
}

@techreport{icao9303,
  title={Doc 9303 -- Machine Readable Travel Documents},
  author={{International Civil Aviation Organization}},
  institution={ICAO},
  year={2021}
}

@standard{iso30107,
  title        = {Information Technology---Biometric Presentation Attack Detection---Part 3: Testing and Reporting},
  organization = {ISO/IEC},
  number       = {30107-3:2017},
  year         = {2017}
}

@inproceedings{blasingame2024greedy,
  title={Greedy-dim: Greedy algorithms for unreasonably effective face morphs},
  author={Blasingame, Zander W and Liu, Chen},
  booktitle={IEEE International Joint Conference on Biometrics (IJCB)},
  pages={1--11},
  year={2024},
}

@inproceedings{damer2018morgan,
  title={MorGAN: Recognition vulnerability and attack detectability of face morphing attacks created by generative adversarial network},
  author={Damer, Naser and Saladie, Alexandra Mosegui and Braun, Andreas and Kuijper, Arjan},
  booktitle={IEEE International Conference on Biometrics Theory, Applications and Systems (BTAS)},
  pages={1--10},
  year={2018},
}

@article{damer2023mordiff,
  title={MorDIFF: Recognition vulnerability and attack detectability of face morphing attacks created by diffusion autoencoders},
  author={Damer, Naser and Fang, Meiling and Siebke, Patrick and Kolf, Jan Niklas and Huber, Marco and Boutros, Fadi},
  journal={arXiv preprint arXiv:2302.01843},
  year={2023}
}

@article{zhang2021mipgan,
  title={{MIPGAN}--Generating strong and high quality morphing attacks using identity prior driven {GAN}},
  author={Zhang, Haoyu and Venkatesh, Sushma and Ramachandra, Raghavendra and Raja, Kiran and Damer, Naser and Busch, Christoph},
  journal={IEEE Transactions on Biometrics, Behavior, and Identity Science},
  volume={3},
  number={3},
  pages={365--383},
  year={2021},
  publisher={IEEE}
}

@inproceedings{damer2022privacy,
  title={Privacy-friendly synthetic data for the development of face morphing attack detectors},
  author={Damer, Naser and L{\'o}pez, C{\'e}sar A. Fontanillo and Fang, Meiling and Spiller, Noémie and Pham, Minh Vu and Boutros, Fadi},
  booktitle={IEEE/CVF Conference on Computer Vision and Pattern Recognition Workshops (CVPRW)},
  pages={1606--1617},
  year={2022}
}

@inproceedings{neto2022orthomad,
  title={{OrthoMAD}: Morphing attack detection through orthogonal identity disentanglement},
  author={Neto, Pedro C. and Gon{\c{c}}alves, Tiago and Huber, Marco and Damer, Naser and Sequeira, Ana F. and Cardoso, Jaime S.},
  booktitle={International Conference of the Biometrics Special Interest Group (BIOSIG)},
  pages={173--181},
  year={2022},
}

@article{caldeira2025madation,
  title={{MADation}: Face morphing attack detection with foundation models},
  author={Caldeira, Eduarda and Ozgur, Guray and Chettaoui, Tahar and Ivanovska, Marija and Peer, Peter and Boutros, Fadi and Struc, Vitomir and Damer, Naser},
  journal={arXiv preprint arXiv:2501.03800},
  note={Accepted at WACV 2025 Workshops},
  year={2025}
}

@inproceedings{paulo2026fd,
  title={{FD-MAD}: Frequency-domain residual analysis for face morphing attack detection},
  author={Paulo, Diogo J and Proen{\c{c}}a, Hugo and Neves, Jo{\~a}o C},
  booktitle={IEEE 20th International Conference on Automatic Face and Gesture Recognition (FG)},
  pages={1--10},
  year={2026},
}

@inproceedings{paulo2026srl,
  title={{SRL-MAD}: Structured Residual Latents for One-Class Morphing Attack Detection},
  author={Paulo, Diogo J and Proen{\c{c}}a, Hugo and Neves, Jo{\~a}o C},
  booktitle={14th International Workshop on Biometrics and Forensics (IWBF)},
  pages={1--6},
  year={2026},
}

@article{ivanovska2026selfmadpp,
  title={{SelfMAD++}: Self-supervised foundation model with local feature enhancement for generalized morphing attack detection},
  author={Ivanovska Preskar, Marija and Todorov, Leon and Peer, Peter and Struc, Vitomir},
  journal={Information Fusion},
  volume={127},
  pages={103921},
  year={2026},
  publisher={Elsevier}
}

@inproceedings{ivanovska2025selfmad,
  title={{SelfMAD}: Enhancing generalization and robustness in morphing attack detection via self-supervised learning},
  author={Ivanovska, Marija and Todorov, Leon and Damer, Naser and Jain, Deepak Kumar and Peer, Peter and Struc, Vitomir},
  booktitle={IEEE International Conference on Automatic Face and Gesture Recognition (FG)},
  year={2025}
}

@inproceedings{mamede2024fairness,
  title={Fairness under cover: Evaluating the impact of occlusions on demographic bias in facial recognition},
  author={Mamede, Rafael M and Neto, Pedro C and Sequeira, Ana F},
  booktitle={European Conference on Computer Vision},
  pages={84--98},
  year={2024},
  organization={Springer}
}

@article{tian2024impact,
  title={Impact of sunglasses on one-to-many facial identification accuracy},
  author={Tian, Sicong and Wu, Haiyu and King, Michael C and Bowyer, Kevin W},
  journal={IEEE Transactions on Biometrics, Behavior, and Identity Science},
  year={2026},
  publisher={IEEE}
}

@inproceedings{huang2020curricularface,
  title={{CurricularFace}: Adaptive curriculum learning loss for deep face recognition},
  author={Huang, Yuge and Wang, Yuhan and Tai, Ying and Liu, Xiaoming and Shen, Pengcheng and Li, Shaoxin and Li, Jilin and Huang, Feiyue},
  booktitle={IEEE/CVF Conference on Computer Vision and Pattern Recognition (CVPR)},
  pages={5901--5910},
  year={2020}
}

@inproceedings{boutros2022elasticface,
  title={{ElasticFace}: Elastic margin loss for deep face recognition},
  author={Boutros, Fadi and Damer, Naser and Kirchbuchner, Florian and Kuijper, Arjan},
  booktitle={IEEE/CVF Conference on Computer Vision and Pattern Recognition Workshops (CVPRW)},
  pages={1578--1587},
  year={2022}
}

@inproceedings{meng2021magface,
  title={{MagFace}: A universal representation for face recognition and quality assessment},
  author={Meng, Qiang and Zhao, Shichao and Huang, Zhida and Zhou, Feng},
  booktitle={IEEE/CVF Conference on Computer Vision and Pattern Recognition (CVPR)},
  pages={14225--14234},
  year={2021}
}

@inproceedings{deng2019arcface,
  title={{ArcFace}: Additive angular margin loss for deep face recognition},
  author={Deng, Jiankang and Guo, Jia and Xue, Niannan and Zafeiriou, Stefanos},
  booktitle={IEEE/CVF Conference on Computer Vision and Pattern Recognition (CVPR)},
  pages={4690--4699},
  year={2019}
}

@techreport{ngan2020frvtmorph,
  title={Face Recognition Vendor Test (FRVT) Part 4: {MORPH} -- Performance of Automated Face Morph Detection},
  author={Ngan, Mei L. and Grother, Patrick J. and Hanaoka, Kayee K. and Kuo, Jason M.},
  institution={National Institute of Standards and Technology (NIST)},
  number={NISTIR 8292},
  year={2020},
  doi={10.6028/NIST.IR.8292}
}

@article{raja2021sotamd,
  title={Morphing attack detection -- database, evaluation platform, and benchmarking},
  author={Raja, Kiran and Ferrara, Matteo and Franco, Annalisa and Spreeuwers, Luuk and Batskos, Ilias and de Wit, Florens and Gomez-Barrero, Marta and Scherhag, Ulrich and Fischer, Daniel and Venkatesh, Sushma and Mohan Singh, Jag and Li, Guoqiang and Bergeron, Lo{\"i}c and Isadskiy, Sergey and Ramachandra, Raghavendra and Rathgeb, Christian and Frings, Dinusha and Seidel, Uwe and Knopjes, Fons and Veldhuis, Raymond N. J. and Maltoni, Davide and Busch, Christoph},
  journal={IEEE Transactions on Information Forensics and Security},
  volume={16},
  pages={4336--4351},
  year={2021},
  publisher={IEEE}
}

@inproceedings{tapia2023simulating,
  title={Simulating print/scan textures for morphing attack detection},
  author={Tapia, Juan E. and Russo, Maximilian and Busch, Christoph},
  booktitle={European Signal Processing Conference (EUSIPCO)},
  year={2023},
}

@article{tapia2025printscan,
  title={Generating automatically print/scan textures for morphing attack detection applications},
  author={Tapia, Juan E. and Russo, Maximilian and Busch, Christoph},
  journal={IEEE Access},
  year={2025},
  doi={10.1109/ACCESS.2025.3555922},
  note={arXiv preprint arXiv:2408.09558}
}

@incollection{patwardhan2025clip,
  title={Empowering morphing attack detection using interpretable image-text foundation model},
  author={Patwardhan, Sushrut and Ramachandra, Raghavendra and Venkatesh, Sushma},
  booktitle={Computer Vision and Image Processing (CVIP 2024), Communications in Computer and Information Science, vol. 2475},
  year={2026},
  publisher={Springer},
}

@inproceedings{clip,
  title={Learning transferable visual models from natural language supervision},
  author={Radford, Alec and Kim, Jong Wook and Hallacy, Chris and Ramesh, Aditya and Goh, Gabriel and Agarwal, Sandhini and Sastry, Girish and Askell, Amanda and Mishkin, Pamela and Clark, Jack and others},
  booktitle={International Conference on Machine Learning},
  pages={8748--8763},
  year={2021},
  organization={PmLR}
}

@article{hu2022lora,
  title={Lora: Low-rank adaptation of large language models.},
  author={Hu, Edward J and Shen, Yelong and Wallis, Phillip and Allen-Zhu, Zeyuan and Li, Yuanzhi and Wang, Shean and Wang, Liang and Chen, Weizhu and others},
  journal={ICLR},
  volume={1},
  number={2},
  pages={3},
  year={2022}
}

@inproceedings{lfw,
  title={Labeled faces in the wild: A database forstudying face recognition in unconstrained environments},
  author={Huang, Gary B and Mattar, Marwan and Berg, Tamara and Learned-Miller, Eric},
  booktitle={Workshop on faces in'Real-Life'Images: detection, alignment, and recognition},
  year={2008}
}

@article{dosovitskiy2020image,
  title={An image is worth 16x16 words: Transformers for image recognition at scale},
  author={Dosovitskiy, Alexey and Beyer, Lucas and Kolesnikov, Alexander and Weissenborn, Dirk and Zhai, Xiaohua and Unterthiner, Thomas and Dehghani, Mostafa and Minderer, Matthias and Heigold, Georg and Gelly, Sylvain and others},
  journal={arXiv preprint arXiv:2010.11929},
  year={2020}
}

@inproceedings{dargaud2023principal,
  title={A principal component analysis-based approach for single morphing attack detection},
  author={Dargaud, Laurine and Ibsen, Mathias and Tapia, Juan and Busch, Christoph},
  booktitle={IEEE/CVF Winter Conference on Applications of Computer Vision Workshops (WACVW)},
  pages={683--692},
  year={2023},
}

@inproceedings{scherhag2017vulnerability,
  author    = {Scherhag, Ulrich and Raghavendra, Ramachandra and Raja, Kiran B. and Gomez-Barrero, Marta and Rathgeb, Christian and Busch, Christoph},
  title     = {On the Vulnerability of Face Recognition Systems Towards Morphed Face Attacks},
  booktitle = {International Workshop on Biometrics and Forensics (IWBF)},
  pages     = {1--6},
  year      = {2017},
  doi       = {10.1109/IWBF.2017.7935088}
}

@inproceedings{raghavendra2017transferable,
  author    = {Raghavendra, Ramachandra and Raja, Kiran B. and Venkatesh, Sushma and Busch, Christoph},
  title     = {Transferable Deep-{CNN} Features for Detecting Digital and Print-Scanned Morphed Face Images},
  booktitle = {IEEE/CVF Conference on Computer Vision and Pattern Recognition Workshops (CVPRW)},
  pages     = {1822--1830},
  year      = {2017},
  doi       = {10.1109/CVPRW.2017.228}
}

@article{ferrara2021printscan,
  author  = {Ferrara, Matteo and Franco, Annalisa and Maltoni, Davide},
  title   = {Face Morphing Detection in the Presence of Printing/Scanning and Heterogeneous Image Sources},
  journal = {IET Biometrics},
  year    = {2021},
  doi     = {10.1049/bme2.12021}
}

@article{neddo2024impact,
  author  = {Neddo, Richard E. and Blasingame, Zander W. and Liu, Chen},
  title   = {The Impact of Print-Scanning in Heterogeneous Morph Evaluation Scenarios},
  journal = {arXiv preprint arXiv:2404.06559},
  year    = {2024}
}

@inproceedings{zhang2024generalized,
  author    = {Zhang, Haoyu and Ramachandra, Raghavendra and Raja, Kiran and Busch, Christoph},
  title     = {Generalized Single-Image-Based Morphing Attack Detection Using Deep Representations from Vision Transformer},
  booktitle = {IEEE/CVF Conference on Computer Vision and Pattern Recognition Workshops (CVPRW)},
  pages     = {1510--1518},
  year      = {2024},
  doi       = {10.1109/CVPRW63382.2024.00158}
}

@article{frll,
  title={Vulnerability analysis of face morphing attacks from landmarks and generative adversarial networks},
  author={Sarkar, Eklavya and Korshunov, Pavel and Colbois, Laurent and Marcel, S{\'e}bastien},
  journal={arXiv preprint arXiv:2012.05344},
  year={2020}
}
}

\end{document}